\documentclass[11pt,a4paper]{article}
\usepackage{authblk}
\usepackage[hyperref]{naaclhlt2019}
\usepackage{times}
\usepackage{latexsym}
\usepackage{booktabs}
\usepackage{amsfonts}
\usepackage{color}
\usepackage{graphicx}\usepackage{amsmath}
\usepackage[boxed]{algorithm2e}
\DeclareMathOperator*{\argmax}{argmax}
\DeclareGraphicsExtensions{.PNG}

\usepackage{url}

\SetCommentSty{mycommfont}
\SetKwInput{KwInput}{Input}                
\SetKwInput{KwOutput}{Output}              
\SetKwInput{KwRequired}{Required}

\aclfinalcopy 

\title{One-Shot Learning for Text-to-SQL Generation}

\renewcommand*{\Affilfont}{\normalsize\normalfont}
\newsavebox\affbox
\author[1]{Dongjun Lee}
\author[1]{Jaesik Yoon}
\author[2]{Jongyun Song}
\author[2]{Sang-gil Lee}
\author[2]{Sungroh Yoon}
\affil[1]{%
{\Affilfont SAP Labs Korea}}
\affil[2]{%
{\Affilfont Electrical and Computer Engineering, Seoul National University}}
\affil[ ]{\textit {\{dongjun.lee01, jaesik.yoon01\}@sap.com, \{coms1580, tkdrlf9202, sryoon\}@snu.ac.kr}}

\date{}
\renewcommand\Affilfont{\itshape\small}

\begin{document}
\maketitle
\begin{abstract}
Most deep learning approaches for text-to-SQL generation are limited to the WikiSQL dataset, which only supports very simple queries. 
Recently, template-based and sequence-to-sequence approaches were proposed to support complex queries, which contain join queries, nested queries, and other types.
However, \citet{improve-text-to-sql} demonstrated that both the approaches lack the ability to generate SQL of unseen templates.
In this paper, we propose a template-based one-shot learning model for the text-to-SQL generation so that the model can generate SQL of an untrained template based on a single example. 
First, we classify the SQL template using the Matching Network \citep{matching-net} that is augmented by our novel architecture Candidate Search Network.
Then, we fill the variable slots in the predicted template using the Pointer Network \citep{pointer-net}.
We show that our model outperforms state-of-the-art approaches for various text-to-SQL datasets in two aspects: 1) the SQL generation accuracy for the trained templates, and 2) the adaptability to the unseen SQL templates based on a single example without any additional training.

\end{abstract}

\section{Introduction}
We focus on a text-to-SQL generation, the task of translating a question in natural language into the corresponding SQL.
Recently, various deep learning approaches have been proposed for the task.
However, most of these approaches target the WikiSQL dataset \citep{wikisql} that only contains very simple and constrained queries \citep{sqlnet, typesql, coarse-to-fine, meta-learning}. \textcolor{black}{These approaches cannot be applied directly to generate complex queries containing elements such as join, group by, and nested queries.}

\citet{improve-text-to-sql} proposed two different approaches to support complex queries: a template-based model and a sequence-to-sequence model. However, both of these models have limitations. The template-based model cannot generate queries of unobserved templates. It requires a lot of examples and additional training to support new templates of SQL. On the other hand, the sequence-to-sequence model is unstable because of the large search space including outputs with SQL syntax errors. Moreover, \citet{improve-text-to-sql} demonstrated that the sequence-to-sequence model also lack the ability to generate SQL queries of unseen templates.

In this work, we propose an extension of a template-based model with one-shot learning, which can generate SQL queries of untrained templates \textit{based on a single example}. Our model works in two phases. The first phase classifies an SQL template. We applied Matching Network \citep{matching-net} for the classification since it is robust to adapt to new SQL templates without additional training. However, as most of the one-shot learning methods, including Matching Network, focus on $n$-way classification setting, it cannot be directly applied to classify a label from a large number of classes. \textcolor{black}{Therefore, we introduce a novel architecture Candidate Search Network that picks \textit{the top-$n$ most relevant SQL templates}. It enables the Matching Network to be utilized to find the most appropriate template among all possible templates.}
\textcolor{black}{The second phase fills the variable slots of the predicted template using a Pointer Network \citep{pointer-net} as these variables are chosen from the tokens in the input sentence.}

The proposed model has three advantages.
\medskip

1. The model is not limited to any particular format of SQL, unlike recent sketch-based approaches \citep{sqlnet, typesql} based on the WikiSQL dataset.
\medskip

2. It minimizes unnecessary search space, unlike sequence-to-sequence approaches \citep{iyer, improve-text-to-sql}; thus, the model is guaranteed to be free of SQL syntax errors.
\medskip

3. The model not only generates SQL of trained templates, but it can also adapt to queries of unseen templates based on a single example \textit{without additional training}.
\medskip

\textcolor{black} {Our approach has great strengths in terms of practical application. To support the SQL queries of new templates, previous approaches require a number of natural language examples for each template and the retraining of the model. In contrast, our model just needs a single example and no retraining.
Moreover, our model is not merely limited to generating SQL but can also be applied to the other code generation tasks \citep{django, hearthstone, nl2bash} by defining templates of code and variables for each template.} 

We conducted experiments with four different text-to-SQL datasets on both of the \textit{question-based split} and \textit{query-based split} \citep{improve-text-to-sql}.
In the question-based split, SQL queries of the same template appear in both training dataset and test dataset. With the question-based split, we tested the effectiveness of the model at generating queries for the trained templates of SQL. 
In contrast, query-based split ensures that queries of the same template only appear in either training or test dataset.
With the query-based split, we studied how well the model can adapt to new templates of SQL.

The experimental result shows that our approach outperforms the state-of-the-art approach by 3--9\% for the question-based split. In addition, we achieved up to 52\% performance gain for the query-based split in the one-shot setting \textit{without additional training}.

\begin{figure*}
	\centering\includegraphics[scale=0.8]{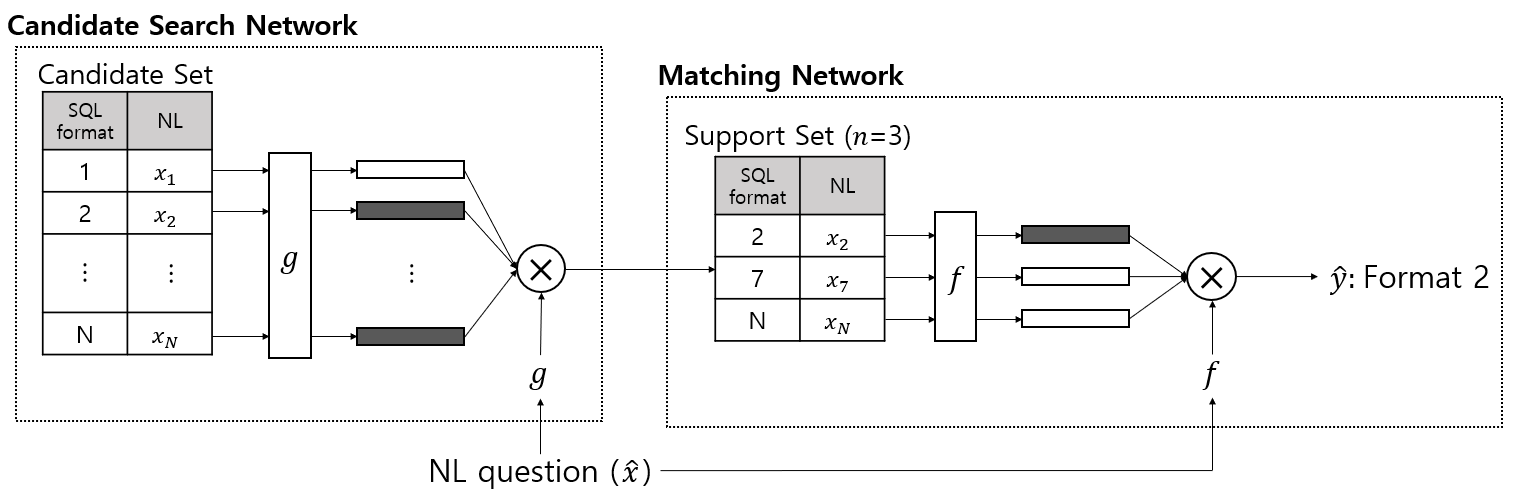}
	\caption{\label{format-clf} The architecture of our SQL template classification model. 
	We propose a Candidate Search Network (CSN) for the selection of top-$n$ relevant SQL templates within the candidate set $\mathcal{C}$ to build a support set $\mathcal{S}$. Then, we find the template $\hat{y}$ using the Matching Network based on the support set $\mathcal{S}$.}
\end{figure*}

\section{Related Work}

{\bf Semantic Parsing}
\textcolor{black}{Semantic parsing is the task of mapping natural language utterances onto machine-understandable representations of meaning. As a sub-task of semantic parsing, natural language to code generation aims to convert a natural language description to the executable code \citep{django, hearthstone, nl2bash}. To solve this task, a variety of deep learning approaches have been proposed. Early works applied a sequence-to-sequence architecture that directly maps a natural language description to a sequence of the target code \citep{ling2016latent, jia2016data}, but this approach does not guarantee syntax correctness. To overcome this limitation, tree-based approaches such as sequence-to-tree \citep{seq2tree} and Abstract Syntax Tree (AST) \citep{ast} have been proposed to ensure syntax correctness. However, \citet{improve-text-to-sql} showed that the sequence-to-tree approach was inefficient when generating complex SQL queries from a natural language question.}

\textcolor{black}{Very recently, \citet{retrieval} proposed a retrieval-based neural code generation (RECODE), sharing a similar idea with our template-based approach. They searched for similar sentences in training dataset using a sentence similarity score and then extracted $n$-grams to build subtrees of AST to be used at the decoding step. In contrast, we introduced an end-to-end learning architecture to retrieve similar sentences in terms of SQL generation. In addition, we do not need a decoding step or subtrees since we generate the full template at once via classification.}
\medskip


{\bf Text-to-SQL}
\textcolor{black}{Natural language interface to database(NLIDB) is a topic that has been actively studied for decades. In early works, there have been several rule-based approaches to parse natural language as SQL \citep{rule1, rule2}. Because rule-based systems suffer from variations in natural language input, enhanced methods leveraging user-interaction have been proposed \citep{rule2, sqlizer}.}

\textcolor{black}{Recently, WikiSQL dataset \citep{wikisql}, a large dataset of natural language and SQL pairs, has been released, and a number of studies have proposed text-to-SQL approaches based on deep learning \citep{sqlnet, typesql, coarse-to-fine, meta-learning}. However, as WikiSQL only contains simple SQL queries, most of the approaches are restricted to the simple queries alone. }

\textcolor{black}{\citet{iyer, improve-text-to-sql} focused on the dataset that contains more complex queries such as ATIS \citep{atis-2} and GeoQuery \citep{geo}.}
To support complex queries, \citet{iyer} applied a sequence-to-sequence approach with attention mechanism, and \citet{improve-text-to-sql} proposed a template-based model and another sequence-to-sequence model with a copy mechanism. However, \citet{improve-text-to-sql} showed that both approaches lack the ability to generate SQL of the unseen template in the training stage.
\medskip



{\bf One-shot Learning/Matching Network}
Deep learning models usually require hundreds or thousands of examples in order to learn a class. To overcome this limitation, one-shot learning aims to learn a class from a single labeled example. \textcolor{black}{We applied one-shot learning to the text-to-SQL task so that our model could learn a SQL template from just a few examples and adapt easily and promptly to the SQL of untrained templates.}

\citet{matching-net} proposed a Matching Network \textcolor{black}{that aims to train an end-to-end $k$-nearest neighbor (kNN) by combining feature extraction and a differentiable distance metric with cosine similarity.}
It enables the model to produce test labels for unobserved classes given only a few samples without any network tuning. 
 However, the $n$-way classification setting used in the Matching Network cannot be directly applied to the general classification problem, because it fixes the number of target classes with a small number $n$ by sampling from whole possible classes.
 Therefore, we introduced a novel architecture Candidate Search Network that chooses the top-$n$ most relevant classes from the entire classes to support the Matching Network. 
\medskip

{\bf Pointer Network}
Pointer Network \citep{pointer-net} aims to predict an output sequence as probability distributions over the tokens in the input sequence. It has been successfully applied to question answering \citep{pointer-question-answering}, abstractive summarization \citep{pointer-summarization}, and code generation \citep{pointer-code-generation}.
We adapted the Pointer Network to fill the variables of the predicted SQL template as these variables are chosen from the tokens in the input sentence.

\section{Approach}
Our approach works in two phases. We first classify an SQL template for a given natural language question and then, we fill the variable slots of the predicted template. 
\textcolor{black}{This architecture is based on an idea similar to the template-based model of \citet{improve-text-to-sql}. However, the previous model requires a number of examples for each template and needs retraining to support new templates of SQL. Conversely, we applied one-shot learning so that our model could learn a template with just a single example. Moreover, our model does not require any additional training to support new SQL templates.}


\subsection{SQL Template Classification}
The SQL template classification model consists of two networks. First, the Candidate Search Network chooses the top-$n$ most relevant templates from a candidate set $\mathcal{C}$ to build a support set $\mathcal{S}$. Then, the Matching Network predicts the SQL template based on the support set $\mathcal{S}$. The overall architecture is depicted in Figure~\ref{format-clf}.
\medskip

{\bf Candidate Search Network}
We propose the Candidate Search Network (CSN) to apply the $n$-way classification setting of the Matching Network to the general classification problem.
First, we build a candidate set $\mathcal{C} = \{(x_i^c, y_i^c)\}_{i=1}^{N}$ which comprises sample pairs of natural language questions and their labels (SQL templates), by sampling one example pair from each of whole classes ($N$) in the training dataset.
For a given test sample $\hat{x}$, the CSN chooses the top-$n$ most relevant items with $\hat{x}$ from the candidate set $\mathcal{C}$ to build a support set $\mathcal{S}=\{ (x_i^s, y_i^s) \}_{i=1}^{n}$. Since the Matching Network assumes that the support set is given, the CSN plays a key role in finding a SQL template among all possible templates via the Matching Network.


To build the CSN, we first trained a convolutional neural network (CNN) text classification model \citep{cnn-text-clf} with the training dataset. From this network, we took features from the last layer before the final classification layer in order to get a feature vector $g(\hat{x})$ and ${\{g(x^c_i)\}}_{i=1}^{N}$. Then, we choose the top-$n$ most similar items with $\hat{x}$, using the cosine similarity of the feature vectors to build a support set $S$ for $\hat{x}$.
\medskip

{\bf Matching Network}
A Matching Network consists of an encoder and an augmented memory. 
Encoder $f(\cdot)$ embeds the natural language question as a fixed-size vector. We used a CNN as our encoder. \textcolor{black}{It consists of different window sizes of convolutional layers and a max-pooling operation is applied over each feature map. The output of the encoder is the concatenated vector of each pooled feature.}

The augmented memory stores a support set $\mathcal{S}$ that is generated by the CSN.
For a given test example $\hat{x}$, our classifier predicts a label $\hat{y}$ based on the support set $\mathcal{S}=\{ (x_i^s, y_i^s) \}_{i=1}^{n}$ as follows:
\begin{equation}
\hat{y} = \sum_{i=1}^n a(\hat{x}, x_i^s) y_i^s
\label{equation-1}
\end{equation}
where $a(\cdot, \cdot)$ is an attention function defined as follows:
\begin{equation}
a(\hat{x}, x_i^s) = e^{c(f(\hat{x}), f(x_i^s))} / \sum_{j=1}^n e^{c(f(\hat{x}), f(x_j^s))}
\end{equation}
where $c(\cdot, \cdot)$ denotes cosine similarity.

For training, we followed the $n$-way 1-shot training strategy. We first sampled label sets $\mathcal{L}$ size of $n$ from all the possible labels $N$. Then we sampled one example for each label in $\mathcal{L}$ to build the support set $\mathcal{S}$. Finally, we sampled a number of examples for each label in $\mathcal{L}$ to build a training batch $T$ to train the model.
The training objective is to maximize the log-likelihood of the predicted labels in batch $T$, based on the support set $\mathcal{S}$ as follows:
\begin{equation}
\argmax_\theta {\mathbb{ E}_{\mathcal{L} \sim N} \left[ \mathbb{E}_{\mathcal{S},T \sim \mathcal{L}} \left[ \sum_{(x, y) \in T} \log p_\theta(y \mid x, \mathcal{S}) \right] \right]} .
\end{equation}

\begin{table*}[h!]
\centering
\begin{tabular}{lcccc}
\toprule
 & \# questions & \# vocabularies & \# SQL templates & avg \# of variables \\
 \midrule
Advising & 4385 & 2371 & 205 & 1.8 \\
ATIS & 5280 & 725 & 947 & 3.4 \\
GeoQuery & 877 & 279 & 246 & 0.7 \\
Scholar & 817 & 716 & 193 & 1.5 \\
\bottomrule
\end{tabular}
\caption{\label{dataset-table} Descriptive statistics of text-to-SQL datasets. }
\end{table*}

\subsection{Slot-Filling}
\begin{figure}[h!]
	\centering\includegraphics[scale=0.38]{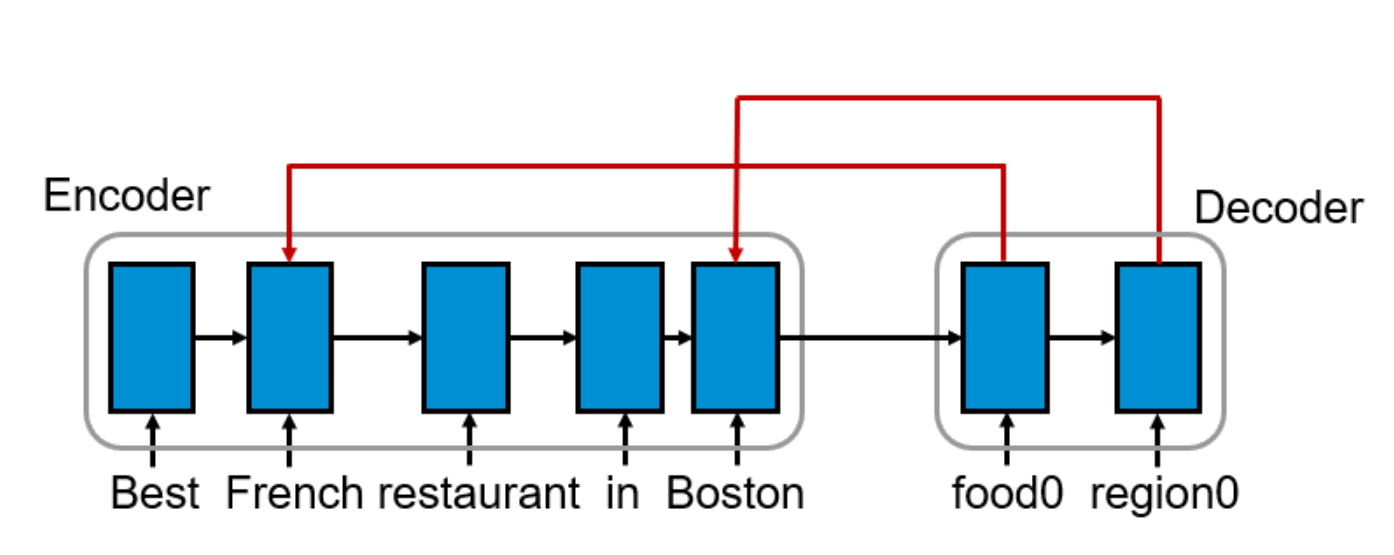}
	\caption{\label{pointer-net} Pointer Network for slot filling. }
\end{figure}

We applied the Pointer Network \citep{pointer-net} to fill the variable slots of the predicted SQL template, as described in Figure~\ref{pointer-net}.

We used a bi-directional LSTM as an input encoder and a uni-directional LSTM as an output decoder. Let $(x_1, ... ,x_n)$ denote the tokens in a natural language question and $(v_1, ..., v_m)$ denote the variables of the SQL template. Then the encoder hidden states are $(e_1, ..., e_n)$, while the decoder hidden states are $(d_1, ..., d_m )$. At each time step $t$ in the decoding phase (for each variable $v_t$), we computed the attention vector as follows:
\begin{equation}
u_i^t = V\; \texttt{tanh} (W_1 e_i + W_2 d_t), \quad  i \in (1, ..., n)
\end{equation}
where $W_1$ and $W_2$ are trainable parameters. Then we applied softmax to obtain the likelihood over the tokens in the input sentence as follows:
\begin{equation}
p(y_t \mid  y_1, ... y_{t-1}, x) = \texttt{softmax} (u^t)
\end{equation}
where $y=(y_1, ..., y_m)$ is a sequence of indices, each between 1 and $n$.

The training objective is to maximize the log-likelihood of the predicted tokens for the given natural language input and list of variables in the SQL template as follows: For the parameter sets $\phi$ of the Pointer Network,
\begin{equation}
    \argmax_\phi {\sum_{(x, v, y) \in D} \log p_\phi(y \mid x, v)}
\end{equation}
where $D$ denotes training dataset.

\subsection{Adaptation} \label{section3.3}
Our model can be adapted to the new template of SQL with a single example, \textit{without additional training}. Assume that there is a natural language to SQL template pair $(x^\prime, y^\prime)$ and that $y^\prime$ is the unseen template during the training. We merely need to add one example pair $(x^\prime, y^\prime)$ to the candidate set $\mathcal{C}$ to make our model applicable to the new template $y^\prime$.

\subsection{Training and Inference}

\begin{algorithm}[h]
\DontPrintSemicolon
  \KwRequired{Candidate set $\mathcal{C}: \{(x_i^c, y_i^c)\}_{i=1}^{N}$ where $N$ is number of SQL templates.}
  \KwRequired{$n:$ hyperparameter for support set size}
  \KwInput{Natural language question $\hat{x}$}
  \KwOutput{SQL template $\hat{y}^{t}$, variables $\hat{y}^{v}$}
  \BlankLine
  $\mathcal{S} : \{ (x_i^s, y_i^s) \}_{i=1}^{n}  \leftarrow CSN(\hat{x}, \mathcal{C})$  \tcp*{Build a support set for $\hat{x}$}
  \BlankLine
  $\hat{y}^{t} \leftarrow MN(\hat{x}, \mathcal{S})$ \tcp*{Predict the template using Matching Network based on the support set $\mathcal{S}$}
  \BlankLine
  $\hat{v} \leftarrow $ list of variables in $\hat{y}^{t}$ \;
  $\hat{y}^{v} \leftarrow PtrNet(\hat{x}, \hat{v})$ \tcp*{Predict the variables in the natural language tokens using Pointer Network.}
  \BlankLine
\caption{\label{inference-algorithm} Inference steps.}
\end{algorithm}

\textcolor{black}{Our approach has three module parameter sets, $\pi$ (CSN), $\theta$ (Matching Network), and $\phi$ (Pointer Network). We train these three modules independently. As our Matching Network uses the same CNN architecture as CSN, we set the initial weight of the Matching Network using trained parameters from CSN for the efficient training.  At the inference stage, we apply the three modules consecutively as described in Algorithm~\ref{inference-algorithm}.}

\section{Experiments}

\subsection{Dataset}

We used four different text-to-SQL datasets for experiments.
\medskip

\textbf{Advising} \citep{improve-text-to-sql} \textcolor{black}{Collection of questions on a course information database at a university. Questions with corresponding SQL were collected from a web page and by students, and augmented by paraphrasing with manual inspection.}
\medskip

\textbf{Atis} \citep{atis-1, atis-2} \textcolor{black}{Collection of questions on a flight booking system. We used a SQL version of the dataset processed by \citet{improve-text-to-sql}.}
\medskip

\textbf{GeoQuery} \citep{geo} \textcolor{black}{Collection of questions on a US geography database. We used a SQL version of the dataset processed by \citet{improve-text-to-sql}.}
\medskip

\textbf{Scholar} \citep{iyer} \textcolor{black}{Collection of questions on an academic publication database. Questions were collected by the crowd, and initial corresponding SQL were automatically generated by the system and augmented with manual inspection.}
\medskip

\textcolor{black}{The number of natural language questions, vocabularies, SQL templates, and the average number of variables per SQL template for each dataset is described in Table~\ref{dataset-table}.}
\textcolor{black}{We used a template and variables for each SQL from the preprocessed versions provided by \citet{improve-text-to-sql}. For the question-based split, we used a 2:1:1 ratio for the train:dev:test split and ensured that every SQL template in the test set appeared at least once in the training set. For the query-based split, we used the same split as in \citet{improve-text-to-sql}.}

\subsection{Model Configuration}
We used the same hyperparameters for every dataset.
For the word embedding, we used deep contextualized word representations (ELMO) from \citet{elmo} (1024 dimensions).
For the encoder of the Matching Network, we used convolution filter window sizes of 2, 3, 4, 5 with 200 feature maps each, and a rectified linear unit(ReLU) as an activation function. We used 15-way 1-shot learning to train the Matching Network; the implication of this is that the Candidate Search Network selects the top-15 most relevant SQL templates. For the Pointer Network, we used a 1-layer 256-unit bi-directional LSTM as an encoder and 1-layer 256-unit uni-directional LSTM as a decoder. For the optimization, we used Adam optimizer \citep{adam} with a learning rate of 0.001 and used early stopping with 100 epochs for a batch size of 64.

\subsection{Experimental Setup}

We evaluated the query generation accuracy for both the \textit{question-based split} and \textit{query-based split} \citep{improve-text-to-sql}. In the question-based split, SQL queries of the same template appear in both train and test sets. Through the question-based split, we tested how well the model could generate SQL of trained templates from natural language questions.
On the contrary, the query-based split ensures that SQL queries of the same template only appear in either train or test set. Through the query-based split, we evaluated how well the model can generalize unseen templates of queries.

\begin{figure}[h!]
	\centering\includegraphics[scale=0.62]{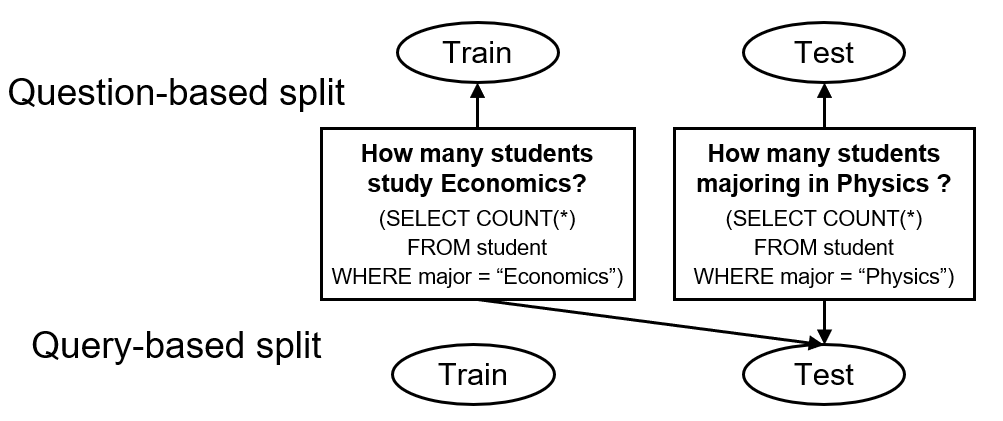}
	\caption{\label{split-figure} Question-based split and query-based split. }
\end{figure}

For the query-based split, we studied our model in two scenarios: \textit{zero-shot} and \textit{one-shot}. In the zero-shot scenario, we trained the model with a training dataset and evaluated it with a test dataset, which is the same setting under previous approaches. In the one-shot scenario, we first trained the model with the training dataset. Then, we sampled a single example from each SQL templates in the test dataset for adaptation. Finally, we evaluated our adapted model with the remaining test dataset. Through the one-shot scenario, we examined how well our model adapts to the unseen templates of SQL from a single example.

\begin{table*}[t!]
\centering
\begin{tabular}{l|c|c|c|c}
\toprule
Model & Advising & ATIS & GeoQuery & Scholar \\
\midrule
Ours & \textbf{89} & \textbf{71} & \textbf{83} & \textbf{67} \\
\midrule
Seq2seq \citep{improve-text-to-sql} & 62 & 49 & 80 & 62 \\
Template \citep{improve-text-to-sql} & 80 & 63 & 78 & 60 \\
\citet{iyer} & 33 & 53 & 68 & 43 \\
\bottomrule
\end{tabular}
\caption{\label{result-table-1} \textcolor{black}{SQL generation accuracy for the question-based split.}
}
\end{table*}

\begin{table*}[t!]
\centering
\begin{tabular}{l|cc|cc|cc|cc}
\toprule
 & \multicolumn{2}{c}{Advising} & \multicolumn{2}{c}{ATIS} & \multicolumn{2}{c}{GeoQuery} & \multicolumn{2}{c}{Scholar} \\
Model & "0" & "1" & "0" & "1" & "0" & "1" & "0" & "1" \\
\midrule
Ours & 0 & \textbf{65} & 0 & \textbf{34} & 0 & \textbf{67} & 0 & \textbf{42} \\
\midrule
Seq2seq \citep{improve-text-to-sql} & 0 & 5 & \textbf{32} & 27 & 20 & 66 & \textbf{5} & 22 \\
Template \citep{improve-text-to-sql} & 0 & 13 & 0 & 17 & 0 & 27 & 0 & 23 \\
\citet{iyer} & \textbf{1} & 3 & 17 & 20 & \textbf{40} & 23 & 3 & 15 \\
\bottomrule
\end{tabular}
\caption{\label{result-table-2} \textcolor{black}{SQL generation accuracy for the query-based split in a zero-shot setting ("0" column) and a one-shot setting ("1" column)}
}
\end{table*}

\subsection{Baselines}
\textcolor{black}{We compare our results with three different previous approaches: a sequence-to-sequence model from \citet{iyer}, template-based model, and another sequence-to-sequence model from \citet{improve-text-to-sql}. \citet{iyer} proposed an encoder-decoder model with global attention \citep{luong} to directly generate a sequence of SQL tokens from a natural language question. \citet{improve-text-to-sql} proposed a template based model using a bi-directional LSTM. The LSTM output for each word was used to predict whether the word is one of the variables or not, and the last hidden state of the LSTM was used to predict the template of SQL. They also proposed a sequence-to-sequence model with attention \citep{bahdanau} and copying mechanism \citep{copy-mechanism} to copy variables in the natural language tokens to the SQL output.}

\textcolor{black}{To test the one-shot setting scenario, because previous approaches cannot perform adaptation without retraining, we added one-shot examples to the training dataset and retrained each of previous model. On the contrary, we did not retrain our model but just added one-shot examples to the candidate set as mentioned in Section~\ref{section3.3}.}

\section{Results and Analysis}

\subsection{Comparison to Previous Approaches}

\textcolor{black}{Table~\ref{result-table-1} shows the results of the query generation accuracy for the question-based split. Table~\ref{result-table-2} shows the query generation accuracy for the query-based split both in a zero-shot setting and a one-shot setting.}
\medskip

{\bf Question-based Split}
For the question-based split, our model outperformed the state-of-the-art approaches in every benchmark. Our model shows 3--27\% query generation accuracy gain, compared to the sequence-to-sequence model, 5--9\% gain, compared to template-based model \citep{improve-text-to-sql}, and 15--56\% gain, compared to \citet{iyer}. \textcolor{black}{The result demonstrates that our model is more efficient in generating SQL of the trained templates than the previous approaches.}
\medskip

{\bf Query-based Split}
Although our approach cannot generate a SQL of unseen templates, we observed that it could adapt well to new templates of SQL given just a single example without additional training.
Sequence-to-sequence models \citep{iyer, improve-text-to-sql}, as shown in the Table~\ref{result-table-2}, showed poor performance for the query-based split in the zero-shot setting. \textcolor{black}{The model from \citet{improve-text-to-sql} showed accuracies of 0\%, 32\%, 20\%, and 5\% for each benchmark and accuracies of \citet{iyer} showed 1\%, 17\%, 40\%, and 3\%, meaning that they also lack the capability to generate unseen templates of SQL.}

\textcolor{black}{In a one-shot setting, where an example is added for each new template, our approach outperformed previous ones against every benchmark. Our model outperforms the sequence-to-sequence model \citep{improve-text-to-sql} by 1--60\%, the template-based model \citep{improve-text-to-sql} by 17--52\%, \citet{iyer} by 14--62\%. It should be noted that previous models were retrained with one-shot examples as they cannot adapt to unseen templates of SQL without additional training. In contrast, we did not retrain our model but merely added one-shot examples to the candidate set in memory. This result demonstrates that our model is considerably more efficient in adapting SQL of new templates than the previous approaches even \textit{in the absence of any additional training.}  }


\begin{table*}[h!]
\centering
\begin{tabular}{l|c|c|c|c}
\toprule
Model & Advising & ATIS & GeoQuery & Scholar \\
\midrule
Our Approach & {\bf 71.2} & {\bf 38.5} & {\bf 67.0} & {\bf 41.8} \\
\quad - Candidate Search Network & 32.0 & 21.3 & 25.0 & 22.8 \\
\quad - Matching Network & 60.8 & 23.2 & 57.1 & 28.5 \\
\bottomrule
\end{tabular}
\caption{\label{mn-csn-result-table} Ablation analysis of our SQL template classification model for \textit{query-based split} in the \textit{one-shot setting}. We report the classification accuracy for each of the following: 1) Our approach (CSN + Matching Network), 2) Only the Matching Network, and 3) Only the CSN. }

\end{table*}

\begin{table*}[h!]
\centering
\begin{tabular}{l|cc|cc|cc|cc}
\toprule
 & \multicolumn{2}{c}{Advising} & \multicolumn{2}{c}{ATIS} & \multicolumn{2}{c}{GeoQuery} & \multicolumn{2}{c}{Scholar} \\
& ? & Q & ? & Q & ? & Q & ? & Q \\
\midrule
SQL template classification & 90 & 71 & 74 & 39 & 83 & 67 & 67 & 42 \\
\quad (Candidate Search Network) & 99 & 93 & 95 & 80 & 100 & 96 & 94 & 75 \\
\quad (Matching Network) & 91 & 76 & 77 & 48 & 83 & 70 & 71 & 56 \\
\midrule
Slot filling & 97 & 91 & 95 & 83 & 100 & 98 & 98 & 93 \\
\bottomrule
\end{tabular}
\caption{\label{breakdown-result-table} \textcolor{black}{Breakdown of results for our approach for both question-based split('?' column) and query-based split('Q' column). For the query-based split, we show the result from the one-shot setting.}}
\end{table*}

\subsection{Ablation Analysis}

\textcolor{black}{To examine how effectively the Candidate Search Network (CSN) and the Matching Network perform together, we conducted an ablation analysis for the \textit{query-based split} in the \textit{one-shot setting}.} The result is shown in the Table~\ref{mn-csn-result-table}. We reported the classification accuracy for each, using: 
\\

1)The Matching Network with Candidate Search Network (CSN) as described in the paper
\\

2) Only the Matching Network that uses a full candidate set as a support set, instead of the $n$-way support set
\\

3) Only the CSN that determines the top-1 most relevant template as a predicted template
\\

By deploying an approach that combines the Matching Network with the CSN, we achieved 19.0\%--42.0\% performance gain, compared to when only the Matching Network was used. \textcolor{black}{This result demonstrates that our proposed CSN plays a key role in enabling the Matching Network to be utilized for classifying templates from a large number of possibilities.}
Compared to when only the CSN was used, we achieved 9.9--15.3\% performance gain.

\subsection{Breakdown Analysis}

\textcolor{black}{We performed a breakdown analysis for both the \textit{question-based split} and the \textit{query-based split with a one-shot setting}.} Table~\ref{breakdown-result-table} shows the accuracy of each modules of our model. Our approach consists of two parts: SQL template classification and slot filling. In addition, our template classification model consists of a Candidate Search Network (CSN) and a Matching Network. For the CSN, accuracy is determined based on the inclusion of the actual label among the $n$ candidates. For the Matching Network, we only report classification accuracy when CSN chooses the $n$ candidates correctly. Regarding the slot filling model, we only count it as correct when all the variables in the template are chosen correctly.

\textcolor{black}{In every benchmark, template classification was a more difficult part than slot filling in both the question-based split and the query-based split. The performance of the slot filling model did not degrade significantly from the question-based to the query-based split (2--12\%). By contrast, the template classification performance dropped by 9--35\%. CSN was able to find the top-15 most relevant templates almost perfectly (94--100\%) in the question-based split but the accuracy dropped by 6--19\% in the query-based split. Finally, the Matching Network showed 71--91\% accuracy in the question-based split and 48--76\% accuracy in the query-based split.}


\section{Conclusion}
In this paper, we proposed a one-shot learning model for the text-to-SQL generation that enables the model to adapt to the new template of SQL based on a single example.
Our approach works in two phases: 1) SQL template classification and 2) slot-filling.
\textcolor{black}{For template classification, we proposed a novel Candidate Search Network that chooses the top-$n$ most relevant SQL templates from the entire templates to build a support set. Subsequently, we applied a Matching Network to classify the template based on the support set. For the slot-filling, we applied a Pointer Network to fill the variable slots of the predicted template. }

\textcolor{black}{We evaluated our model in two aspects. We tested the SQL generation accuracy for the trained templates with \textit{question-based split} and the adaptability to the SQL of new templates with \textit{query-based split}. }
\textcolor{black}{Experimental results showed that our approach outperforms state-of-the-art models for the question-based split.} \textcolor{black}{In addition, we demonstrated that our model could efficiently  generate SQL of untrained templates from a single example, without any additional training.}


\bibliography{naaclhlt2019}
\bibliographystyle{acl_natbib}

\newpage





\end{document}